\documentclass[sigconf]{acmart}
\AtBeginDocument{%
  \providecommand\BibTeX{{%
    \normalfont B\kern-0.5em{\scshape i\kern-0.25em b}\kern-0.8em\TeX}}}

\setcopyright{none}
\renewcommand\footnotetextcopyrightpermission[1]{} 

\usepackage[switch]{lineno}
\usepackage{multirow}
\usepackage{array}
\usepackage{bbm}
\usepackage{dutchcal}
\usepackage{color}

\begin{document}

\title{Video Frame Interpolation with Flow Transformer}

\author{Pan Gao}
\affiliation{%
  \institution{Nanjing University of Aeronautics and Astronautics}
  \city{Nanjing}
  \country{P. R. China}}
  \email{Pan.Gao@nuaa.edu.cn}

  \author{Haoyue Tian}
\affiliation{%
  \institution{Nanjing University of Aeronautics and Astronautics}
  \city{Nanjing}
  \country{P. R. China}}
  \email{tianhy@nuaa.edu.cn}

  \author{Jie Qin}
  \authornote{Corresponding author}
\affiliation{%
  \institution{Nanjing University of Aeronautics and Astronautics}
  \city{Nanjing}
  \country{P. R. China}}
  \email{qinjiebuaa@gmail.com}


\renewcommand{\shortauthors}{Gao et al.}

\begin{abstract}
  Video frame interpolation has been actively studied with the development of convolutional neural networks. However, due to the intrinsic limitations of kernel weight sharing in convolution, the interpolated frame generated by it may lose details. In contrast, the attention mechanism in Transformer can better distinguish the contribution of each pixel, and it can also capture long-range pixel dependencies, which provides great potential for video interpolation. Nevertheless, the original Transformer is commonly used for 2D images; how to develop a Transformer-based framework with consideration of temporal self-attention for video frame interpolation remains an open issue. In this paper, we propose Video Frame Interpolation Flow Transformer to incorporate motion dynamics from optical flows into the self-attention mechanism. Specifically, we design a Flow Transformer Block that calculates the temporal self-attention in a matched local area with the guidance of flow, making our framework suitable for interpolating frames with large motion while maintaining reasonably low complexity. In addition, we construct a multi-scale architecture to account for multi-scale motion, further improving the overall performance. Extensive experiments on three benchmarks demonstrate that the proposed method can generate interpolated frames with better visual quality than state-of-the-art methods.
\end{abstract}





\maketitle
\pagestyle{plain}

\section{Introduction}
Video frame interpolation (VFI) is an actively studied problem in the field of computer vision, with a wide range of potential applications in video post-processing \cite{zhao2019cnn}, video restoration \cite{xu2019quadratic,wang2019edvr}, and slow motion video generation \cite{paliwal2020deep,peleg2019net,niklaus2018context}. VFI aims to increase the frame rate of video sequences \cite{meyer2015phase,yu2019hierarchical}, by calculating a frame that does not exist between consecutive input frames, making the video smooth enough and reducing motion blur.

Most previous studies using deep learning for VFI are based on convolutional neural networks (CNNs). These methods either synthesize the interpolated frame after aligning the reference frame by predicting the optical flow \cite{xue2019video,liu2019deep,niklaus2020softmax}, or directly perform convolution operations on the reference frame by estimating a spatially adaptive kernel \cite{liu2017video,jiang2018super,bao2019memc,bao2019depth}.

Despite achieving promising results on standard datasets, due to the translation invariance and weight sharing characteristics of convolution \cite{krizhevsky2017imagenet}, the above methods still has some problems in the interpolation process. Specifically, CNNs usually adopt small convolution kernels to extract spatial information at adjacent positions, failing to cope with large-motion scenes; even stacking multiple convolution layers has limited improvement \cite{wang2018non}. In addition, during convolution, different locations in the image share the same kernel parameters, which may cause artifacts and motion blur in the interpolated frame \cite{shi2022video}.

\begin{figure}[t]
    \centering
    \includegraphics[width=0.9 \linewidth]{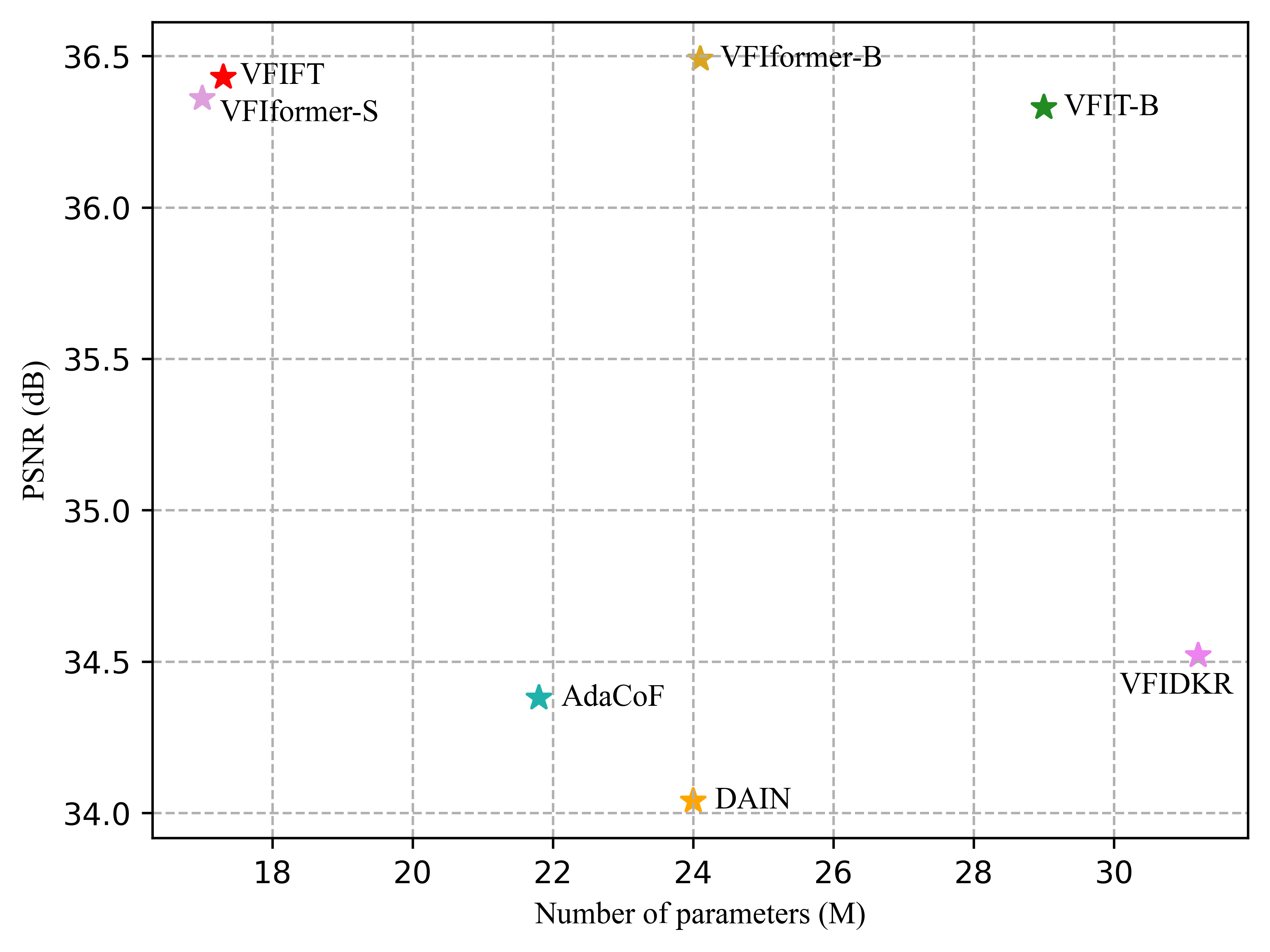}
    
    \caption{Comparison of the performance and parameters of some excellent models under the Vimeo90K dataset.  VFIFT represents our proposed Transformer-based model, which outperforms state-of-the-art CNN-based methods with less parameters. Compared to recent Transformer models, VFIFT still demonstrates very competitive performance with improved synthesis performance or similar performance but having significantly less parameters.}
    \label{chart}
		
\end{figure}

Therefore, a couple of very recent works \cite{shi2022video,lu2022video} began to exploit Transformers for VFI, owing to the long-range dependency modeling capability of Transformers  \cite{dosovitskiy2020image}. For instance, \cite{shi2022video} proposed VFIT that avoids the same weight coefficients and considers large-scale motion scenes. Moreover, to avoid the high computational cost of global self-attention, it extends local attention to the spatio-temporal domain.
However, the temporal self-attention in VFIT is conducted on the temporal vector that is formulated using the co-located pixels across frames, which neglected the motion dynamics of the objects in the video. As a result, the co-located pixel in the reference frame may not exactly correspond to the target pixel, leading to sub-optimal interpolation performance.
To better capture the motion information, VFIformer \cite{lu2022video} introduces optical flow into their Transformer-based framework, as optical flow can well represent temporal dependencies and has been widely used in CNN-based VFI models. However, VFIformer only simply feeds the flow to the Transformer block, which cannot fully benefit from the advantages of optical flow for modeling motion dynamics. In other words, how to incorporate optical flow to Transformer-based VFI models still remains an open yet challenging problem.

\begin{figure*}[t]
    \centering
    \includegraphics[width=0.8 \linewidth]{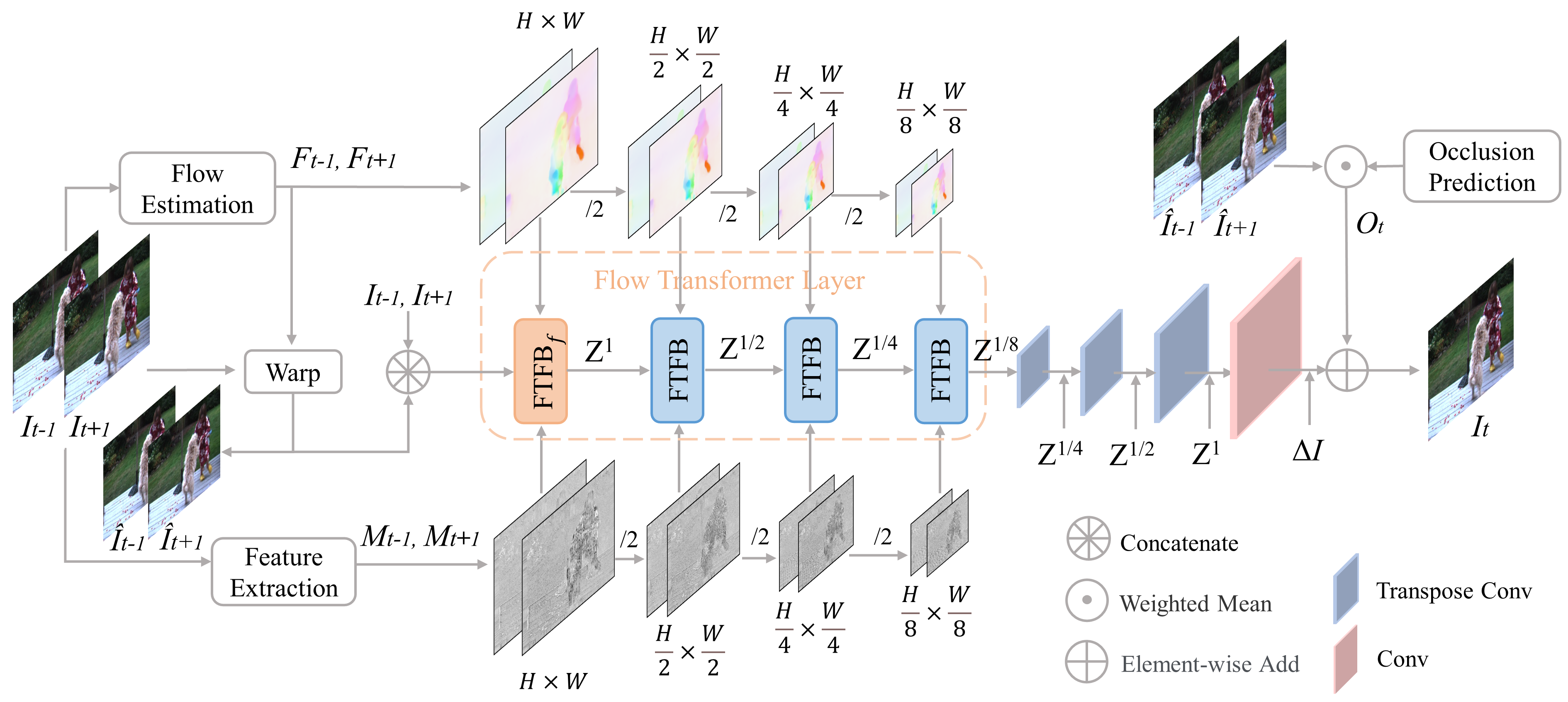}
    
    \caption{The overall architecture of the proposed network. The input of the model consists of two consecutive frames $I_{t-1}$ and $I_{t+1}$, and the optical flow and feature map are obtained through the optical flow prediction network and the feature extraction network respectively. Then, the specific region of the feature maps is located according to the optical flow at multi-scale resolutions, and after that, the model performs the flow attention. The intermediate results are upsampled 3 times and output residual $\Delta I$ of the same size as the input frames. Finally, the blended frame $O_t$ after occlusion processing is added with the residual to get the target frame $I_t$.}
    \label{overview}
    
\end{figure*}

To fully unleash the potential of optical flow in Transformer-based VFI, we present a Video Frame Interpolation Flow Transformer (\emph{i.e.}, VFIFT) by delicately injecting optical flow into the temporal self-attention calculation. In particular, VFIFT first captures the motion information of the target frame relative to the reference frames through optical flow prediction. Based on the motion information, our model locates the corresponding reference area on feature maps, from which the temporal self-attention is finally calculated. Since our proposed VFIFT calculates the self-attention in the local area pointed by the flow, it can better model motion dynamics in video interpolation while maintaining a low computational complexity. The main contributions of this paper are as follows:

\begin{itemize}
    \item We proposed a novel Transformer-based VFI framework, in which optical flow directly guides the temporal self-attention calculation. The proposed VFIFT can boost the interpolation performance while maintaining a reasonable low computational complexity.
    \item We build up a multi-scale Transformer architecture using multi-scale flow, which can enhance the capability of modeling multi-scale motions in diverse videos. 
    \item Our model is evaluated on three typical benchmarks, where the experimental results quantitatively and qualitatively demonstrate the superiority of VFIFT.
\end{itemize}

Figure \ref{chart} shows the comparison between our method and other cutting-edge competitors on the Vimeo90K dataset \cite{xue2019video}. As can be observed, our VFIFT significantly surpasses state-of-the-art CNN-based methods with less parameters. Even for the most advanced Transformer-based methods (\emph{i.e.}, VFIT and VFIformer), our model respectively achieves 0.10dB and 0.07dB improvement in PSNR, with approximately the same number of parameters.

\section{Related Work}

\subsection{Video Frame Interpolation}

The flow-based method predicts the motion information from the reference frame to the target frame by using deep learning and then calculates the value of each target pixel. But it inevitably produce ghosting or blurring artifacts when the input frames are not well aligned. In addition, bilinear interpolation is used when calculating target pixels, which leads to less relevant reference to spatial information (only $2 \times 2$).
	
 In order to extract more spatial information and reduce the dependence on pixel-wise optical flow, Niklaus \emph{et al.} proposed AdaConv \cite{niklaus2017video1} and Sepconv \cite{niklaus2017video2} model to estimate the spatial-adaptive convolutional kernels of each output pixel, that is, the target pixel value is obtained by convolution on the local patch of the reference frames. Besides, AdaCoF \cite{lee2020adacof} generated the output frame by estimating the kernel weight and offset vector of each target pixel.

For better use of optical flow information and kernel information, MEMC \cite{bao2019memc} first aligns the input frame through the optical flow information, and then calculates the interpolated pixels through local patches. DAIN \cite{bao2019depth} also exploits depth information to clearly detect occlusions for video frame interpolation. \cite{tian2022video} also provides a corresponding offset for each reference pixel via a neural network so that it can better cover the object itself and improve the quality of the output image.

\subsection{Vision Transformer}

In natural language processing, Transformer has always been in an important position. 
Although previous work attempted to replace convolutions in vision architectures, it was only exploded recently when Dosovitisky showed pure Transformer networks with their ViT \cite{dosovitskiy2020image} architecture also achieve state-of-the-art results for image classification. After that, Liu \emph{et al.} proposed a new backbone, called Swin Transformer \cite{liu2021swin}, which reduces computational complexity by computing attention in local windows. It also proposed a shift window scheme to establish the relationship between non-overlapping windows.

ViViT \cite{arnab2021vivit} used pure Transformer architectures for video classification tasks. It extracted spatial-temporal tokens from the input video, which are then encoded by a series of Transformer layers. Some variants of their model are proposed, including those that are more effective by decomposing the spatial and temporal dimensions of the input video.

VFIT \cite{shi2022video} proposed a Transformer-based video interpolation framework, which considered the relationship between the long-term dependence of self-attention and video frame interpolation, and avoided high computing costs through the window attention mechanism of Swin Transformer. However, this method does not  use motion information, but instead seeks dependencies on target pixels across the entire feature map. Concurrently, Lu \emph{et al.} \cite{lu2022video} addressed video frame interpolation with Transformer, i.e., the aforementioned VFIformer, where they first align the reference frames via the optical flow, and then put the warped frames into the model. Nevertheless, this approach is very sensitive to the accuracy of the optical flow prediction. If the optical flow prediction is not accurate, this will lead to incorrect warped frames, which will undoubtedly directly affect subsequent pixel generation.

Inspired by the above studies, we propose a novel motion-aware Transformer structure, specifically design  a cross-frame window-based self-attention
mechanism to achieve state-of-the-art interpolation performance. We introduce the VFIFT to combine the motion information with the attention mechanism. This module makes full use of optical flow to accurately locate each point to the corresponding reference area on the input frame, which greatly improves the image quality and increases the efficiency of our model.

\section{Method}

The purpose of video frame interpolation is to synthesize the target frame $I_t$ that does not exist in-between through the adjacent reference frames. Figure \ref{overview} shows the overall structure of our proposed method, where it is mainly divided into three steps, namely information extraction, flow transformer layer and occlusion prediction. We aim to generate intermediate frame $I_t$ with two consecutive frames $I_{t-1}$ and $I_{t+1}$. The following sections will introduce these three steps in detail and analyze the improvement in time efficiency of our approach.

\subsection{Information Extraction}

\begin{figure}[t]
    \centering
    \includegraphics[width=0.8 \linewidth]{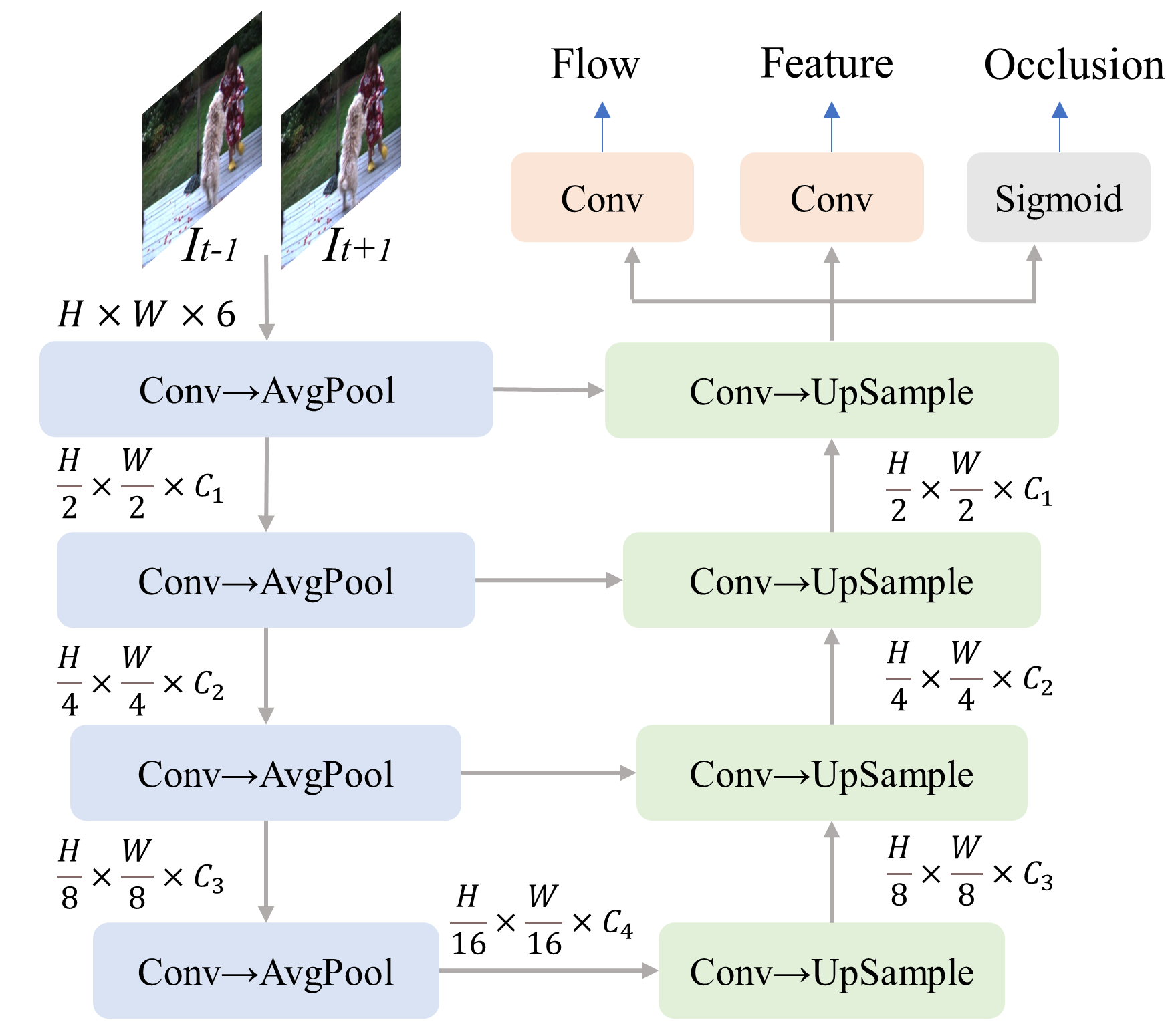}
    
    \caption{The structure diagram designed for flow estimation, feature extraction and occlusion prediction. We input the reference images into the network, through the process of downsample and upsample, which enables the model to learn at multiple resolutions, and finally output different results to represent optical flow, feature map and occlusion coefficient depending on the last layer.}

    \label{network}
\end{figure}

Considering that the optical flow plays a guiding role in the alignment of the target frame, we first use the optical flow to warp the input frames, and then calculate the attention in the warped frames. Therefore, we design a flow prediction network to obtain the motion trajectories of the target frame relative to the forward and backward reference frames, respectively. The network structure is shown in Figure \ref{network}.

We use the U-Net \cite{ronneberger2015u} structure as the basic framework of the flow prediction network, which is also the framework of the feature extraction network and the occlusion prediction network. Specifically, it first downsamples the input frame 4 times, which is achieved by average pooling, and then upsamples the feature map by the same number of times using transposed convolution for reconstruction. Note that the feature map for each upsampling is the concatenation of the previous layer output and the feature map with the same resolution in the downsampling stage.

We design the last layer of the model differently to reach different functionalities of the three modules. For the flow prediction network, it finally passes through a convolutional layer and outputs two flow maps of $H \times W \times 2$, which respectively represent the motion vector ($F_{t-1}$ and $F_{t+1}$) between the target frame and the input frames, where $2$ in the dimension refers to the $x$ direction and the $y$ direction.
The last layer of the feature extraction network is also a convolutional layer, which outputs two feature maps ($M_{t-1}$ and $M_{t+1}$) of $H \times W \times C$. They respectively represent the features extracted from two input frames, where $C$ is a hyper-parameter we need to set. For the occlusion prediction network, the difference is that the output will eventually pass through the sigmoid activation function to obtain a visibility coefficient map with a shape of $H \times W \times 1$, and its value is between $[0, 1]$, which can be understood as the reference weight for the warped frame $\hat{I}_{t-1}$.

\subsection{Flow Transformer Layer}

We designed our main component as shown in Figure \ref{FTFB}(a), which is called Flow TransFormer Block (FTFB). Specifically, first this block concatenates the input with the feature map of the corresponding resolution and downsamples it by setting the stride of the convolution to $2$. Then we perform a $3 \times 3$ convolution and use LeakyReLU \cite{xu2015empirical} as the activation function. The next is to enter the proposed Flow TransFormer Attention (FTFA) module, the details of which are shown in Figure \ref{FTFB}(b). The features in the FTFA are processed as:
\begin{align}
	\hat{k}^o = FA(LN(k^{i})) + k^{i},  \\
	k^o = MLP(LN(\hat{k}^o)) + \hat{k}^o
\end{align}
where $k^i$ and $k^o$ represent the input and output of FTFA, LN and MLP denote the LayerNorm and Multilayer Perceptron, respectively. FA represents the Flow Attention in Figure \ref{FTFB}(b), which uses optical flow ($F_{t-1}, F_{t+1}$) to calculate local attention, it will be described in detail later. Finally, the output is obtained through $3 \times 3$ convolution. Note that FTFB$_f$ is the first FTFB module in the Flow Transformer Layer, and there are two differences between FTFB$_f$ in Figure \ref{overview} and this module. To be specific, the input received by FTFB$_f$ are the input frames ($I_{t-1}, I_{t+1}$) and warped frames ($\hat{I}_{t-1}, \hat{I}_{t+1}$), while the input received by FTFB is the output of the previous block. Besides, FTFB$_f$ does not perform downsample. 

Although convolution has some inherent limitations, which is the sharing of kernel parameters that may not distinguish the contribution of different reference points, its locality is still very suitable for video and image tasks. Inspired by this, we thus apply local attention. On the one hand, it pays more attention to the target object itself, avoiding the influence of some irrelevant factors such as background information or other objects, on the other hand, local attention can also improve computational efficiency. Furthermore, considering the dependence of the video frame interpolation task on motion information, we also introduce optical flow into our method.

\begin{figure}[t]
    \centering
    \includegraphics[width=1.0 \linewidth]{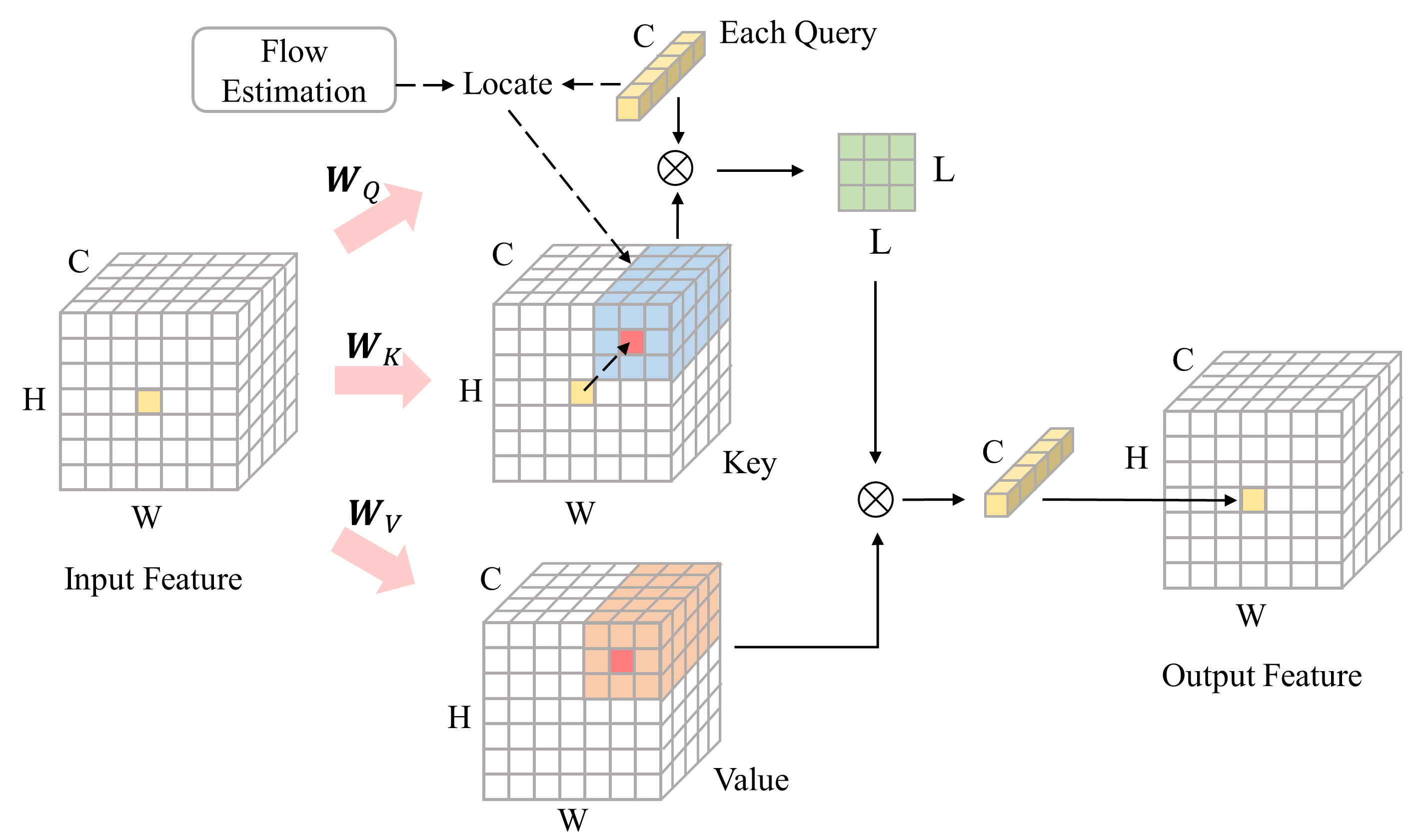}
    
    \caption{The specific process of flow attention. For each query, locate the reference area on the corresponding key and value according to the flow, and perform local attention in it to obtain the result.}
    \label{flowAttention}
\end{figure}

The specific steps are shown in the Figure \ref{flowAttention}. 
We first obtain $Q$, $K$, and $V$ by linearly projecting the input feature, which can be described as:
\begin{equation}
    Q = XW_Q, \quad K = XW_K, \quad V = XW_V,
\end{equation}
where $X \in \mathbb{R}^{H \times W \times C}$ is input feature, $W_Q$, $W_K$ and $W_V$ are the projection matrices. Afterward, we locate the reference area of the current position $(i, j)$ in the $K$ matrix and $V$ matrix through the corresponding optical flow. Since the value of the flow is fractional, we find the closest integer pixel $(i', j')$ by means of the rounding rule. With $(i', j')$ as the center, we find the nearest $L \times L$ size as the local attention range, which means for each pixel query, we only need to perform the attention within this local area. Note that when $(i', j')$ is at the boundary, and this point cannot be the center of the reference area, and if $(i', j')$ is beyond the range of the matrix, we take the point on the matrix closest to it as a new target $(i', j')$. We will elaborate this in the supplementary material. This can be defined as the following:

\begin{figure}[t]
    \centering
    \includegraphics[width=1.0 \linewidth]{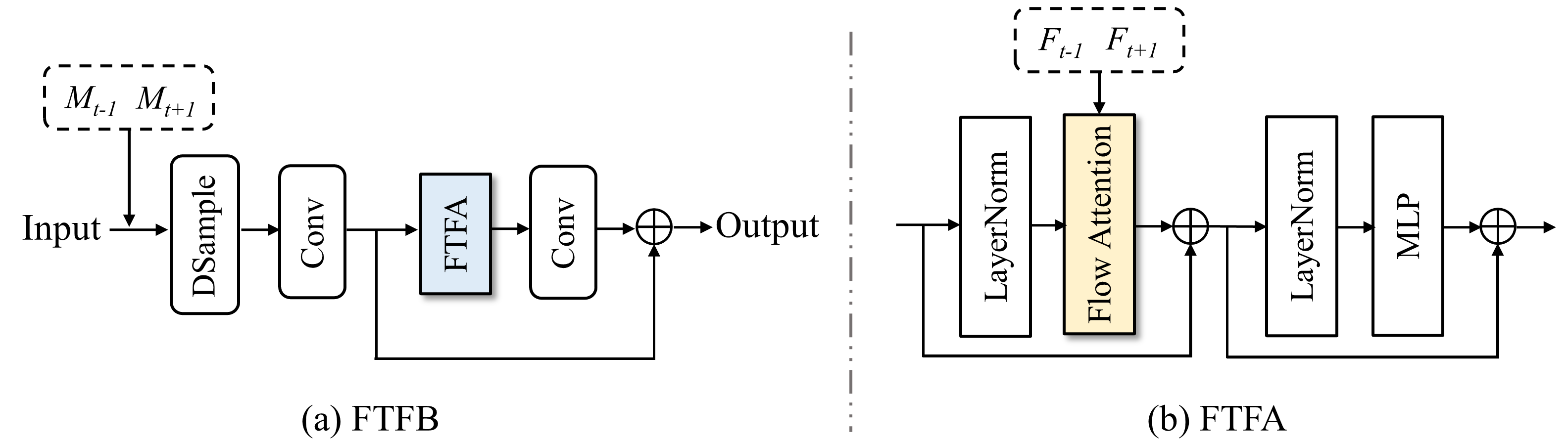}
    
    \vspace{-2mm}
    \caption{(a) represents Flow TransFormer Block (FTFB), and (b) is the datails of the Flow TransFormer Attention (FTFA). }
    
    \vspace{-4mm}
    \label{FTFB}
\end{figure}


\begin{equation}
    FA(X_{(i, j)}) = Softmax \left( \frac{Q_{(i, j)} K_{\rho(i', j')}}{\sqrt{d_k}} \right) \cdot V_{\rho(i', j')}
\end{equation}
where $d_k$ is the key dimension, $X_{(i, j)}$ represents the position of the coordinate point in the input feature, $K_{\rho(i', j')}$ and $V_{\rho(i', j')}$ represent the features at the reference area of size $L \times L$ in the $K$ matrix and the $V$ matrix respectively, and $L$ is usually set to an odd number. Note that the proposed flow attention module employs the flow $F_{t-1}$ to guide the local attention calculation for the input feature from $M_{t-1}$, while using  $F_{t+1}$ for the embedded feature from $M_{t+1}$.

\subsection{Occlusion Prediction}

Moving objects in natural videos often appear occluded in input frames. When this happen, the object becomes invisible in one of the input frames. To address this issue, we use the occlusion prediction network to learn the occlusion maps, which can be understood as the visibility weight for a certain frame. The network structure of occlusion prediction is shown in Figure \ref{network}. We add sigmoid activation at the end of the network, so that the values of the occlusion map ($O_{t-1}$) output by the network are in the range of $[0, 1]$, where $O_{t-1}$ represents the visibility weight for $\hat{I}_{t-1}$. Similarly, the visibility for $\hat{I}_{t+1}$ is marked as $O_{t+1}$, which is calculated by $\mathbbm{1} - O_{t-1}$. The blended frame $O_t$ is generated by:
\begin{equation}
    O_t = O_{t-1} \odot \hat{I}_{t-1} + O_{t+1} \odot \hat{I}_{t+1}
\end{equation}
where $\odot$ is element-wise product. Finally, we obtain a residual image $\Delta I_t$, and adds it with the blended frame $O_t$ to synthesize the final target frame $I_t$. We improve network performance by learning the residual of interpolated frames.

\subsection{Complexity Analysis}

Compared with convolution and self-attention, local attention has obvious computational efficiency improvement. We show the complexity analysis in this subsection and study their advantages and disadvantages.

Given an input feature map of shape $H \times W \times C$, for convolution, assuming that the input channel count is the same as the output, both the length and width of the convolution kernel are $L$, and the resolution before and after the convolution is consistent, we can derive that its time complexity is $\mathcal{O} (HWC^2L^2)$.

As for self-attention mechanism, for clarity, we only analyze the complexity of single head attention, and the computation for the complexity of multi-head is similar. First, it calculates three matrices of $QKV$, respectively. Assuming that the shapes of their projection matrices are $C \times C$, the time complexity of calculating each matrix is $\mathcal{O} (HWC^2)$. In addition, according to the operation rules of attention, the core matrix calculation is $(QK^T)V$, and its complexity is $\mathcal{O} (2H^2W^2C)$. Therefore, the total computational cost is $\mathcal{O} (3HWC^2 + 2H^2W^2C)$. 

The Swin Transformer divides the queries, keys and values into $\frac{H}{L} \times \frac{W}{L}$ windows of size $L \times L \times C$, where $L$ denotes the width and height of window, and then performs self attention on each window, which makes the time complexity $\mathcal{O} (2HWCL^2)$. Adding the time devoted to calculate $QKV$ as before, the total cost is $\mathcal{O} (3HWC^2 + 2HWCL^2)$.

Similar to the Swin Transformer, our proposed flow attention is also based on the local attention. Each query $Q_{(i, j)}$ is located in an area of size $L \times L$ through the optical flow, and perform the core calculation with the key $K_{\rho(i', j')}$ and value $V_{\rho(i', j')}$ in this area. So its time complexity is $\mathcal{O} (3HWC^2 + 2HWCL^2)$ like the Swin Transformer.

Comparing the time complexity of our proposed method with that of convolution, it is found after simplification:
\begin{equation}
    3C + 2L^2 \leq CL^2,  \quad  if \ C >= 3 \ and \ L >= 3
\end{equation}
where $C$ and $L$ are integers. This shows that when the channel of the input feature map is greater than $3$ and the area of local attention is greater than $3\times3$, the complexity of the former is smaller than that of the latter, which becomes more obvious as $C$ and $L$ increase. If compared with self-attention, we can get
\begin{equation}
    L^2 < HW,  \quad  if \ L < H \ and \ L < W
\end{equation}
Since $L$ is the width and height of local area, it is much smaller than the input width and height. So the computational efficiency of local attention is much higher than that of self-attention. As for Swin Transformer, although it has the same complexity as our method in calculating attention, Swin needs to do an additional attention after shifting the window, whose time consumption is thus also larger than our method.

\section{Experiments}

\subsection{Datasets}

\noindent
\textbf{Vimeo90k} \cite{xue2019video}. We use its triplet dataset, which is designed for time frame interpolation and consists of 73,171 3-frame sequences with a fixed resolution of $448 \times 256$. We utilize the first frame and the third frame as the input of the model, and the second frame as the ground truth to compare with the model output.

\noindent
\textbf{UCF101} \cite{soomro2012ucf101}. UCF101 is a real action video dataset with 101 action categories collected from YouTube. It has diversity in action, and there are distinct differences in the shape and pose of objects, complex background and lighting conditions. 
We selected 333 triples for testing, where each frame has a resolution of $320 \times 240$. 

\noindent
\textbf{Davis} \cite{soomro2012ucf101}. This is one of the most important datasets in the video object segmentation task.
We selected 50 groups of three consecutive frames as the test set, each with a resolution of $854 \times 480$. The purpose is to detect the adaptability of the model to harsh environments such as object occlusion, motion blur, and appearance changes.

\subsection{Implementation Details}

For the network structure of flow prediction, feature extraction and occlusion prediction, only the last layer is different, and the U-net framework is used in the main body, where $C_1$, $C_2$, $C_3$ and $C_4$ are set to 32, 64, 128 and 256, respectively. In the FTFB$_f$ and FTFB modules, the size of the local attention is set to $5 \times 5$, and a multi-head attention mechanism with 3 heads is adopted. In addition, the channel number of the feature map output by the flow attention layer and the feature map output after upsampling is set to 160.

The total loss of the proposed model is designed into two parts: the first part is to calculate the loss between the ground truth $I_{gt}$ and the warped frames $\hat{I}_{t-1}$ and $\hat{I}_{t+1}$ synthesized by optical flow, called the warped loss $L_{warp}$. The second part is to calculate the difference of the ground truth $I_{gt}$ and the target interpolated frame $I_t$ output by proposed model, which is called reconstructed loss $L_{rec}$. So the total loss function can be expressed as:
\begin{gather}
    L = \lambda_1 L_{warp} + \lambda_2 L_{rec}  
\end{gather}
where $\lambda_1$ and $\lambda_2$ are the weight coefficients, which are set to 1 and 0.5, respectively. $L_{warp}$ and $L_{rec}$ are obtained by the following formulas:
\begin{gather}
    L_{warp} = \left \| I_{gt} - \frac{\hat{I}_{t-1} + \hat{I}_{t+1}}{2}  \right \|_1   \\
    L_{rec} = \left \| I_{gt} - I_t \right \|_1
\end{gather}

Since the warped frames we get are generated from two reference frames, we average them, and then compare the averaged result with the ground truth. We adopt $\mathbcal{l}_1$ loss as it avoids blurry results in most image synthesis tasks \cite{lee2020adacof}.

We adopt the AdamW \cite{loshchilov2017decoupled} optimizer, where $\beta_1$ and $\beta_2$ are set to default values of 0.9 and 0.999, respectively. In addition, the learning rate is set to 1e-4. We split Viemo90k into two parts for training and testing our proposed model, respectively. One part contains 64,600 triplets as the training set, and the other part contains 8,571 triplets as the test set, with a spatial resolution of 448 × 256 per frame. We take the middle frame of the triplet as ground truth and the remaining two frames as input data. To improve the performance of our model, we randomly crop $192 \times 192$ patches from the training samples and augment them by random flipping and temporal reversal. During training, we set the batch size to 4 and deploy our experiments to RTX 3090. After training for about 200 epochs, the training loss has converged.

\begin{table}[t]
	\centering
 	\caption{Quantitative comparisons on the three datasets under different sizes of reference areas.}
	\renewcommand{\arraystretch}{1.2}
	\begin{tabular}{c|c|c|c|c}
		\toprule[1pt]
		
		& & $7 \times 7$ & $5 \times 5$ & $3 \times 3$ \\
		\midrule
		\multirow{2}{*}{Vimeo90K} & PSNR & 36.32 & \textbf{36.43} & 36.39   \\
		\cline{2-5}
		& SSIM & 0.9805 & \textbf{0.9813} & 0.9812 \\
		\midrule
		\multirow{2}{*}{UCF101} & PSNR & 35.6 & \textbf{35.71} & 35.7  \\
		\cline{2-5}
		& SSIM & 0.9792 & \textbf{0.9795} & \textbf{0.9795} \\
		\midrule
		\multirow{2}{*}{Davis480p} & PSNR & 28.06 & \textbf{28.27} & 28.21  \\
		\cline{2-5}
		& SSIM & 0.8907 & \textbf{0.8915} & 0.8913 \\
		
		\bottomrule[1pt]
	\end{tabular}

	\label{tab1}
\end{table}

\begin{table}[t]
	\centering
 	\caption{Quantitative comparisons on the three datasets under the scenario of using standard convolution to replace the attention mechanism in FTFA.}
	\renewcommand{\arraystretch}{1.2}
	\begin{tabular}{c|c|c|c|c}
		\toprule[1pt]
		
		& & $7 \times 7$ & $5 \times 5$ & $3 \times 3$ \\
		\midrule
		\multirow{2}{*}{Vimeo90K} & PSNR & \textbf{35.72} & \textbf{35.72} & 35.68   \\
		\cline{2-5}
		& SSIM & \textbf{0.9786} & \textbf{0.9786} & 0.9784 \\
		\midrule
		\multirow{2}{*}{UCF101} & PSNR & 35.58 & \textbf{35.6} & 35.51  \\
		\cline{2-5}
		& SSIM & 0.9693 & \textbf{0.9694} & 0.969 \\
		\midrule
		\multirow{2}{*}{Davis480p} & PSNR & \textbf{27.76} & 27.68 & 27.5  \\
		\cline{2-5}
		& SSIM & \textbf{0.8862} & 0.886 & 0.8857 \\
		
		\bottomrule[1pt]
	\end{tabular}

	\label{tab2}
\end{table}

\begin{table}[t]
	\centering
 	\caption{Quantitative comparisons on the three datasets without optical flow.}
	\renewcommand{\arraystretch}{1.2}
	\begin{tabular}{c|c|c|c|c}
		\toprule[1pt]
		
		& & $7 \times 7$ & $5 \times 5$ & $3 \times 3$ \\
		\midrule
		\multirow{2}{*}{Vimeo90K} & PSNR & \textbf{35.64} & 35.6 & 35.59   \\
		\cline{2-5}
		& SSIM & \textbf{0.9788} & 0.9786 & 0.9784 \\
		\midrule
		\multirow{2}{*}{UCF101} & PSNR & \textbf{35.5} & 35.46 & 35.47  \\
		\cline{2-5}
		& SSIM & \textbf{0.979} & 0.9789 & 0.9789 \\
		\midrule
		\multirow{2}{*}{Davis480p} & PSNR & \textbf{27.33} & 27.22 & 27.15  \\
		\cline{2-5}
		& SSIM & \textbf{0.8716} & 0.8711 & 0.8709 \\
		
		\bottomrule[1pt]
	\end{tabular}

	\label{tab3}
	
\end{table}

\begin{table}[tbp]
        \centering
        \caption{PSNR comparison under the three datasets, VFIFT represents the complete structure of proposed network, and VFIFT-D represents the structure after removing the occlusion prediction module. }
	\begin{tabular}{cccc}
		\toprule
		 Methods & Vimeo90k & UCF101 & Davis480p \\
		\midrule
            VFIFT-D & 36.07 & 34.89 & 27.96 \\
		  VFIFT & 36.43 & 35.71 & 28.27 \\
		\bottomrule
		
	\end{tabular}

	\label{ttt2}
\end{table}

\subsection{Ablation Study}
In this section, we conduct several ablation studies on the proposed method. We first verify the performance of the model under different sizes of reference areas, then discuss the effectiveness of the Transformer compared to standard convolutions, and finally analyze the impact of optical flow information on the synthesis of interpolated frames.

\noindent
\textbf{Reference areas}. In experiments, we trained models with reference areas of $7 \times 7$, $5 \times 5$ and $3 \times 3$ respectively, and compared them on the test set. Note that the scope of local attention is not set too large, because the model has been located to an approximate position using optical flow, and the model can obtain good results as long as it searches in small surrounding range. In addition, this can significantly reduce the computational complexity of local attention.

As shown in the Table \ref{tab1}, we found that when the size is set to $5 \times 5$, the overall performance of our model is optimal, and the reference area being too large (such as $7 \times 7$) may contain background information or non-target objects, which will lead to a degradation in the quality of the output image.

\begin{table*}[htbp]
	\centering
 	\caption{Quantitative comparisons on the Vimeo90K, UCF101 and Davis480p. Note that bold data represent the best results, and underlined data indicate second best results. Our Transformer-based model VFIFT can outperform other methods with the same amount of parameters.}
	\renewcommand{\arraystretch}{1.2}
	\begin{tabular}{c|c|c|c|c|c|c|c|c}
		\hline
		
		\multirow{2}{*}{Method} & \multirow{2}{*}{Reference} & \multirow{2}{*}{\#Parameter (M)} & \multicolumn{2}{c|}{Vimeo90K} & \multicolumn{2}{c|}{UCF101} & \multicolumn{2}{c}{Davis480p} \\
		\cline{4-9}
		   &  & & PSNR & SSIM & PSNR & SSIM & PSNR & SSIM \\
		\hline
		MEMC & TPAMI'19 & 67.2 & 34.20 & 0.959 & 35.16 & 0.9632 & 27.27 & 0.8121 \\
		DAIN & CVPR'19 & 24 & 34.04 & 0.9581 & 35.26 & 0.963 & 27.31 & 0.8148 \\
		AdaCoF & CVPR'20 & 21.8 & 34.38 & 0.9562 & 33.93 & 0.9421 & 26.72 & 0.7687 \\
		VFIDKR & IJCAI'22 & 31.2 & 34.52 & 0.9612 & 35.5 & 0.9647 & 27.46 & 0.8164 \\
		VFIformer-S & CVPR'22 & 17.0 & 36.36 & 0.9809 & \underline{35.66} & 0.9792 & 28.13 & 0.8895 \\
		VFIformer-B & CVPR'22 & 24.1 &  \textbf{36.49} & \textbf{0.9814} & 35.43 & 0.97 &  \textbf{28.31} & \underline{0.8912} \\
		VFIT-S & CVPR'22 & 7.5 & 35.96 & 0.9760 & 34.46 & 0.9705 & 27.92 & 0.8859 \\
		VFIT-B & CVPR'22 & 29 & 36.33 & 0.9811 & 34.84 & 0.9709 & 28.09 & 0.8896 \\
		\textbf{VFIFT} & Ours (full) & 17.3 & \underline{36.43} & \underline{0.9813} &  \textbf{35.71} &  \textbf{0.9795} & \underline{28.27} &  \textbf{0.8913} \\
  		\textbf{VFIFT-Conv} & Ours (w/ Conv) & 17.2 & 36.02 & 0.9798 & 35.65 & \underline{0.9793} & 27.9 & 0.8862 \\
		
		\hline
	\end{tabular}

	\label{tab4}
\end{table*}

\begin{table}[tbp]
        \centering
        \caption{We measure the average run time in seconds, required by various models to generate each frame on the Vimeo90k test set. The model is equipped on an RTX 3090 GPU.}
	\begin{tabular}{cccc|c}
		\toprule
		 MEMC & VFIDKR & VFIformer-S & VFIT-S & VFIFT \\
		\midrule
		 0.53 & 0.41 & 0.18 & 0.24 & 0.11 \\
		\bottomrule
		
	\end{tabular}
	\label{ttt1}
\end{table}

\begin{figure*}[!htbp]
    \centering
    \includegraphics[width=0.95 \linewidth]{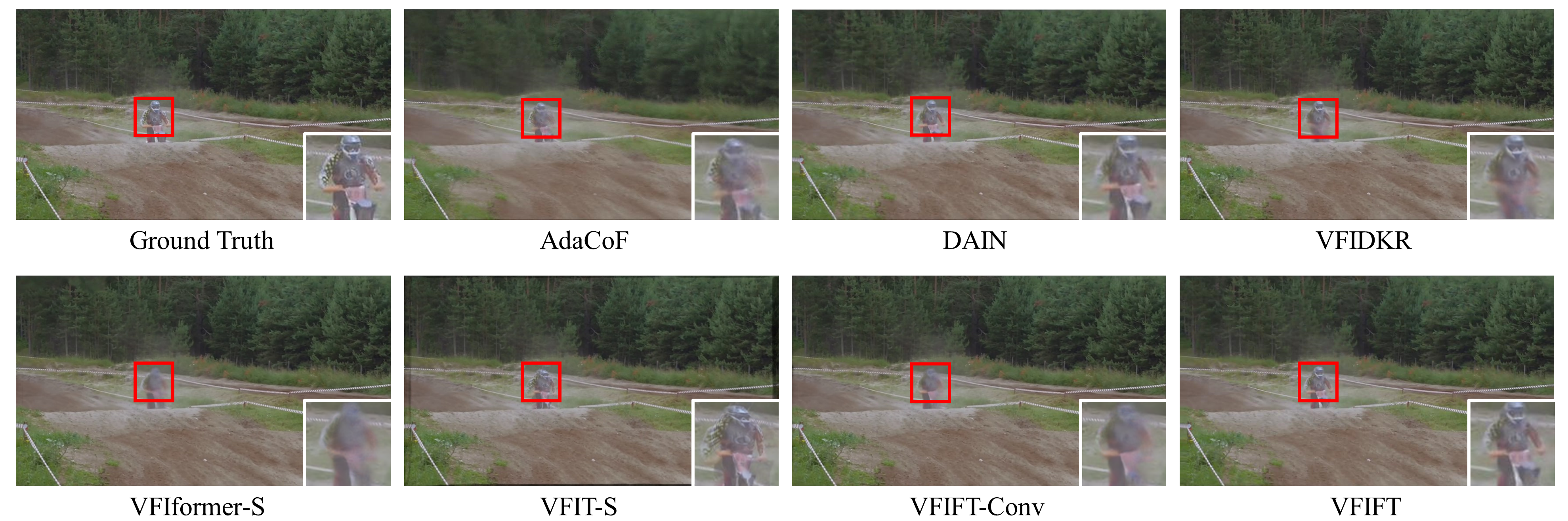}

\end{figure*}

\begin{figure*}[!htbp]
    \centering
    \includegraphics[width=0.95 \linewidth]{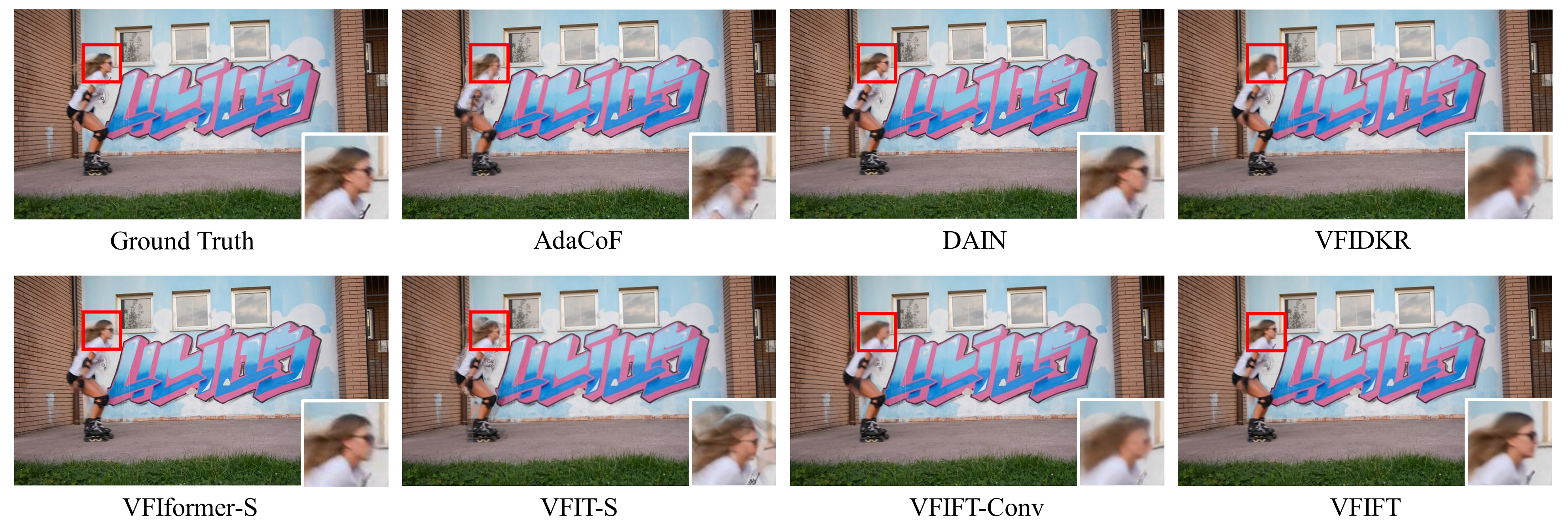}
    \caption{Visual comparison of different methods on the Davis480p test set.}
    
    \label{result}
\end{figure*}

\noindent
\textbf{Attention}. Considering that most of the previous interpolation approaches are based on the CNN model, we compared the results generated by the two different model architectures of Attention and CNN. Specifically, we partially modified the FTFA in our model. The first step, like flow attention, located the general reference area through optical flow, and then do standard convolution operation instead of Attention. For the sake of fairness, we kept the difference on the parameters of these two models within $0.1$M to avoid bias caused by different model scales.

Similarly, the experiment trained the models under three receptive fields of $7 \times 7$, $5 \times 5$ and $3 \times 3$, and the results are shown in the Table \ref{tab2}. By comparing this table with Table \ref{tab1}, we find that under three datasets, the PSNR and the SSIM of the interpolated frame calculated by Transformer are higher than those generated by the standard convolution, which shows that our method can break the limitation of CNN convolution kernel weight sharing.

\noindent
\textbf{Optical flow}. In order to prove the important role of optical flow in synthesizing interpolated frames, we also trained the model without optical flow. The input of the FTFB$_f$ module and the FTFB module is kept unchanged, except when the flow attention step is passed, the model did not locate the reference area through the motion information, but directly performed the attention in the $7 \times 7$, $5 \times 5$ and $3 \times 3$ range around the target point, and the boundary problem is solved in the same way as our proposed scheme. The results are as shown in the Table \ref{tab3} and we find the importance of optical flow localization. In the case of using flow to determine the reference area, the average PSNR of the three data sets can be improved, which is especially obvious for the larger motion under the Davis480p data set.

\noindent
\textbf{Occlusion prediction}. We also did the ablation study on occlusion prediction module. Table \ref{ttt2} shows the quantitative results after we remove the occlusion prediction part in the network framework, which shows that the occlusion prediction module can improve the image quality of the interpolated frames.

\subsection{Comparison with State-of-the-Art Methods}

We compared our model with several recent competing methods, including MEMC \cite{bao2019memc}, DAIN \cite{bao2019depth}, AdaCoF \cite{lee2020adacof}, VFIDKR \cite{tian2022video}, VFIformer-S \cite{lu2022video}, VFIformer-B \cite{lu2022video}, VFIT-S \cite{shi2022video} and VFIT-B \cite{shi2022video}. Table \ref{tab4} shows the quantitative comparison, where the best and second best results are marked in bold and underline, respectively. Among them, VFIFT represents the Transformer-based method we proposed, and VFIFT-Conv is an alternative of our proposed model that replaces Transformer with convolution in FTFA. It is observed that our method is superior to other most advanced methods on the three test sets, except that it is slightly inferior to the results of VFIformer-B under Vimeo90k and Davis480p. This may be because VFIformer-B has 6.8M more parameters than our method. The comparison of visual results between our method and other methods is shown in Figure \ref{result}. It can be observed that for fast-moving objects, the images generated by our method can be smoother without large artifacts. More qualitative results are presented in the supplementary materials.

We also show the running time of our method in Table \ref{ttt1}. Whether compared with CNN-based methods or transformer-based methods, the running time of our proposed VFIFT is smaller than them, which further confirms the time complexity analysis in our paper.

\section{Conclusion}
This paper proposes a new framework for video frame interpolation task. We use Transformer as the basic structure instead of CNN to break the restriction of its convolution kernel weight sharing. In particular, before the attention operation, we introduced the optical flow to roughly locate the reference area, which can make the model more focused on the object itself and adapt well to large motion scenes. Considering the multi-scale motion characteristics of video, we also design a multi-scale flow Transformer architecture. Experiments show that our method performs better than most of the existing advanced methods on multiple test sets. Furthermore, our model has the fastest inference speed.

\begin{acks}
This work was supported by  the Natural Science Foundation of China (No. 62272227 and  No. 62276129), and  the Natural Science Foundation of Jiangsu Province (No. BK20220890).
\end{acks}

\bibliographystyle{ACM-Reference-Format}
\bibliography{ijcai23}



\appendix
\section{Boundary Problems}
	
\begin{figure}[!h]
		\centering
		\includegraphics[width=1.0 \linewidth]{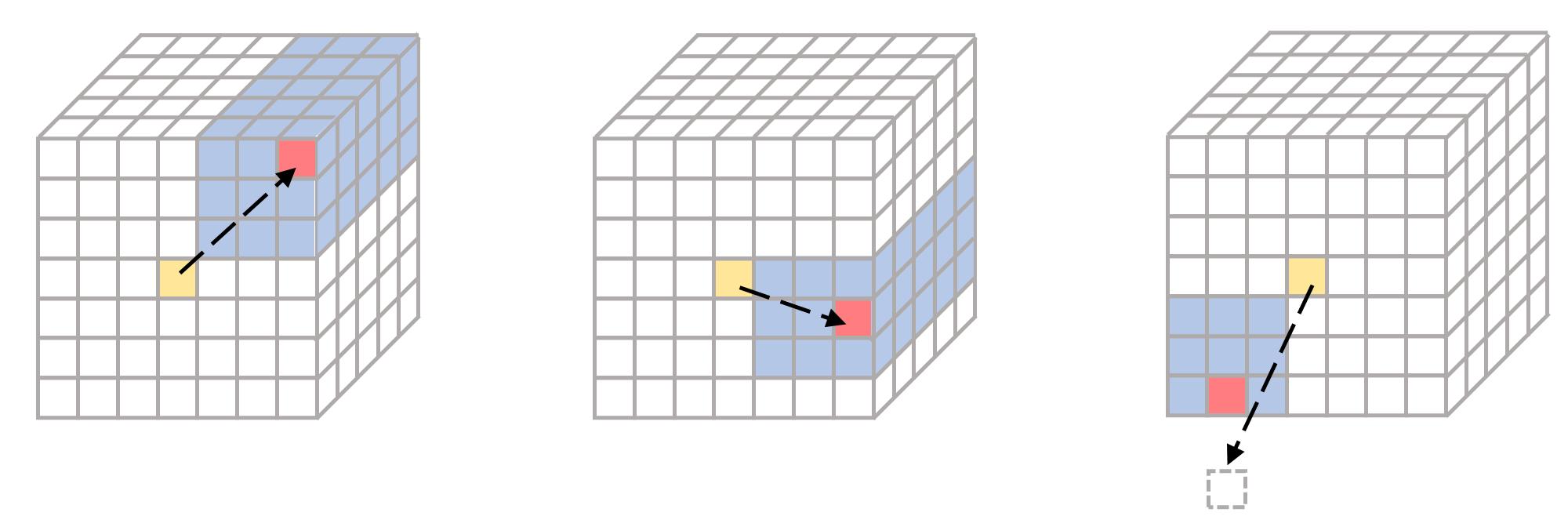}
		
		\caption{Schematic diagram of boundary problems. The left and middle figures represent the processing method of $(i', j')$ at the boundary, and the right figure represents the processing method for the case that point is located outside the matrix range. The yellow point is the position of the target point $(i, j)$, the black dotted arrow is the optical flow, and the red point represents the position of the integer point $(i', j')$ after positioning according to the optical flow. The blue background denotes the reference area $\rho(i', j')$. Here we assume that $L$ is 3.}
		\label{sup1}
	\end{figure}

	In the process of flow attention, moving objects near the boundary may face various local attention overflowing, and we have properly dealt with these problems. When calculating the attention, the model first locates the sub-pixel position of the current target point $(i, j)$ on the matrix K according to the optical flow (note that the matrix V is processed in the same way as the matrix K), and finds the nearest integer pixel position $(i', j')$, as shown by the red point in Figure \ref{sup1}. In general, the model takes $(i', j')$ as the center, finds the nearest $L \times L$ as the reference area $\rho(i',j')$, and perform local attention in it. If $(i', j')$ is coincidentally located at the boundary, as shown in the left and middle of Figure \ref{sup1}, then $(i', j')$ is no longer regarded as the center. But we still make the local attention in the $L \times L$ area closest to this point. Besides, if the point located according to the optical flow exceeds the range of the matrix, we find a matrix point closest to it as new $(i', j')$, and then continue with the method described above, as shown on the right side of Figure \ref{sup1}.
	
	\section{Setting of Ablation Study}
	
\begin{figure}[!h]
		\centering
		\includegraphics[width=1.0 \linewidth]{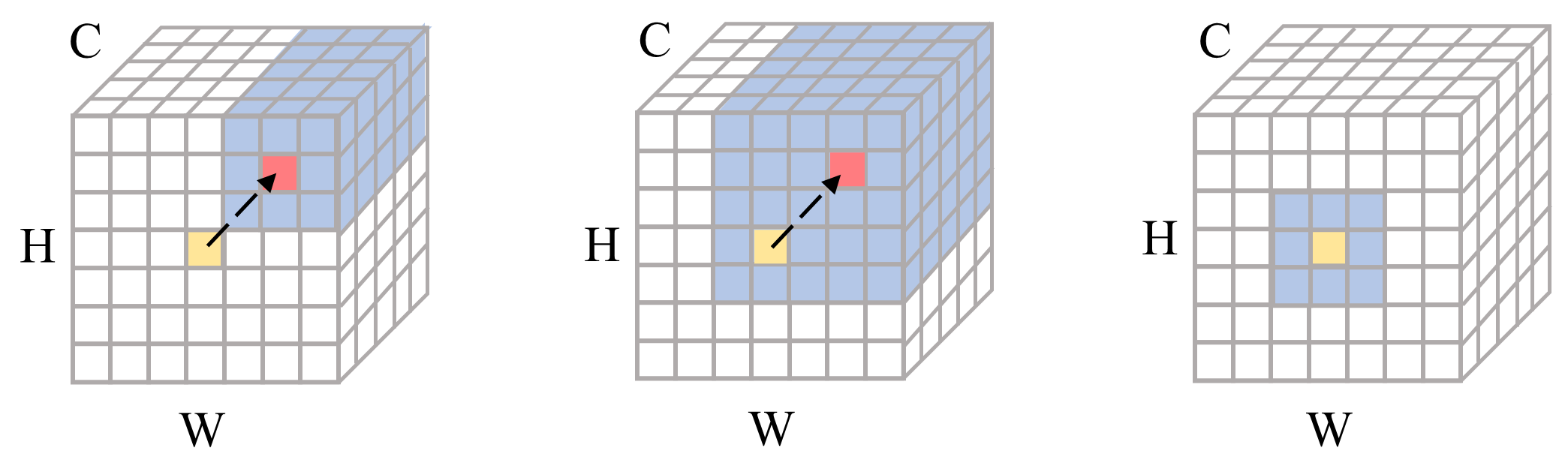}
		
		\caption{Comparison of ablation experiments. The yellow point indicates the position of the current query, the black dotted line represents the optical flow, the red point is the position after being located by the flow, and the blue area indicates the range of the local attention. The figure on the left shows our proposed scheme, the middle one shows the ablation experiment with attention size as the variable, and the right figure shows the study without optical flow.}
		\label{sup10}
	\end{figure}

	We show the difference in ablation experiments by Figure \ref{sup10}. Here we only show the calculation process of the specific query (yellow point) on the $K$ matrix, and the $V$ matrix is similar. The left figure shows the method proposed in this paper, that is, first locate the position of reference point (red point) through optical flow (black dotted line), and then make local attention in the $L \times L$ range (blue area) around it. The figure in the middle represents the process of testing different attention sizes. Besides, the right figure focuses on the influence of optical flow. It does not use any motion information to locate the reference point, but directly uses the current position as the target, and performs local attention directly in the surrounding $L \times L$.

	\section{Visualization of Experimental Results}
	
\begin{figure*}[t]
		\centering
		\vspace{-2mm}
		\includegraphics[width=0.9 \linewidth]{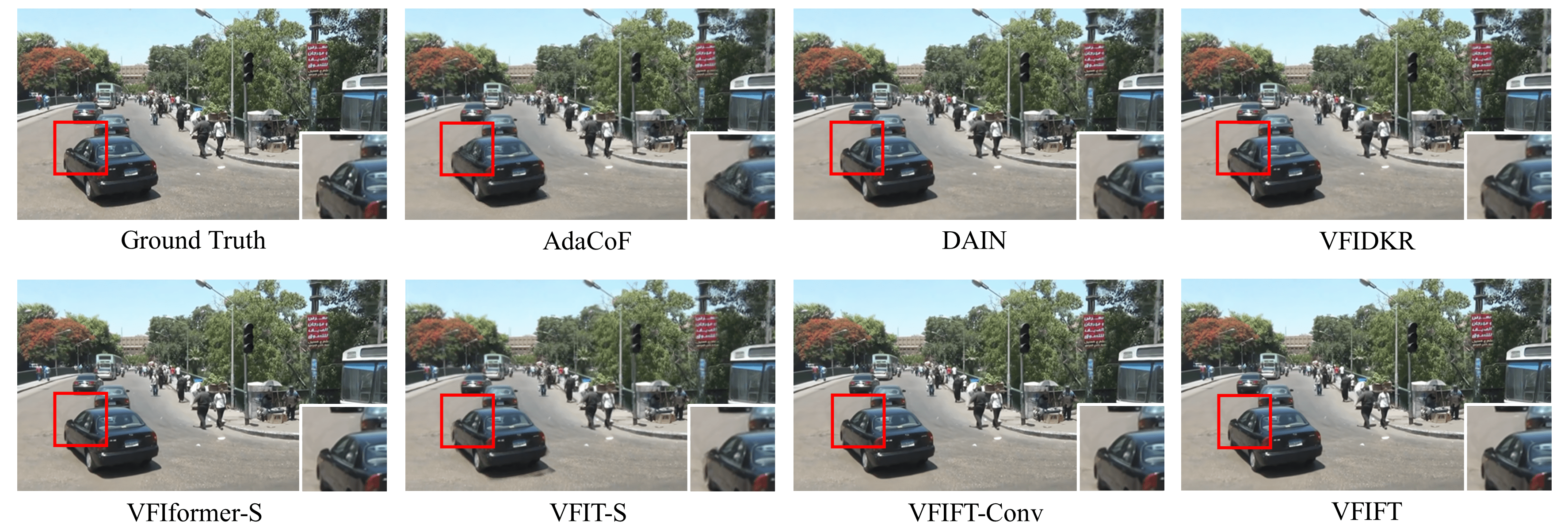}
		
		\caption{Visual comparisons of our proposed method with various state-of-the-art methods on the Vimeo90K.}
		\vspace{-2mm}
		\label{sup2}
	\end{figure*}

\begin{figure*}[t]
		\centering
		\vspace{-2mm}
		\includegraphics[width=0.905 \linewidth]{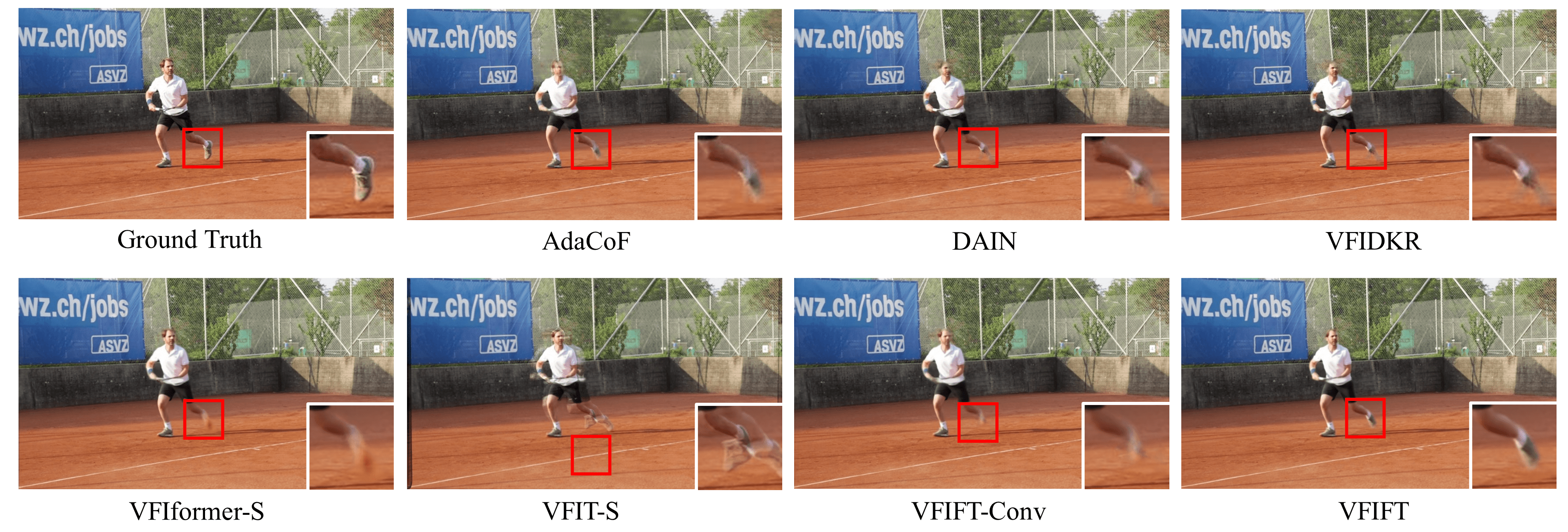}
		
		\caption{Visual comparisons of our proposed method with various state-of-the-art methods on the Davis480p.}
		\vspace{-2mm}
		\label{sup3}
	\end{figure*}

	In order to better prove the effectiveness of the video frame interpolation model proposed in this paper, we provide more qualitative results in this section. We selected several typical sequences from the Vimeo90k and Davis480p datasets for testing, and displayed the interpolated frames to compare the difference in image visual quality between our proposed method and the latest most advanced CNN method and Transformer method, as shown in Figures \ref{sup2} and Figure \ref{sup3}. Through observation, we find that the interpolated frames generated by our method have better subjective image quality, especially for video sequences with complex large motion. In addition, our proposed approach with flow attention block performs visually better than its CNN variant.

	\begin{figure*}[t]
		\centering
		\vspace{-5mm}
		\includegraphics[width=0.9 \linewidth]{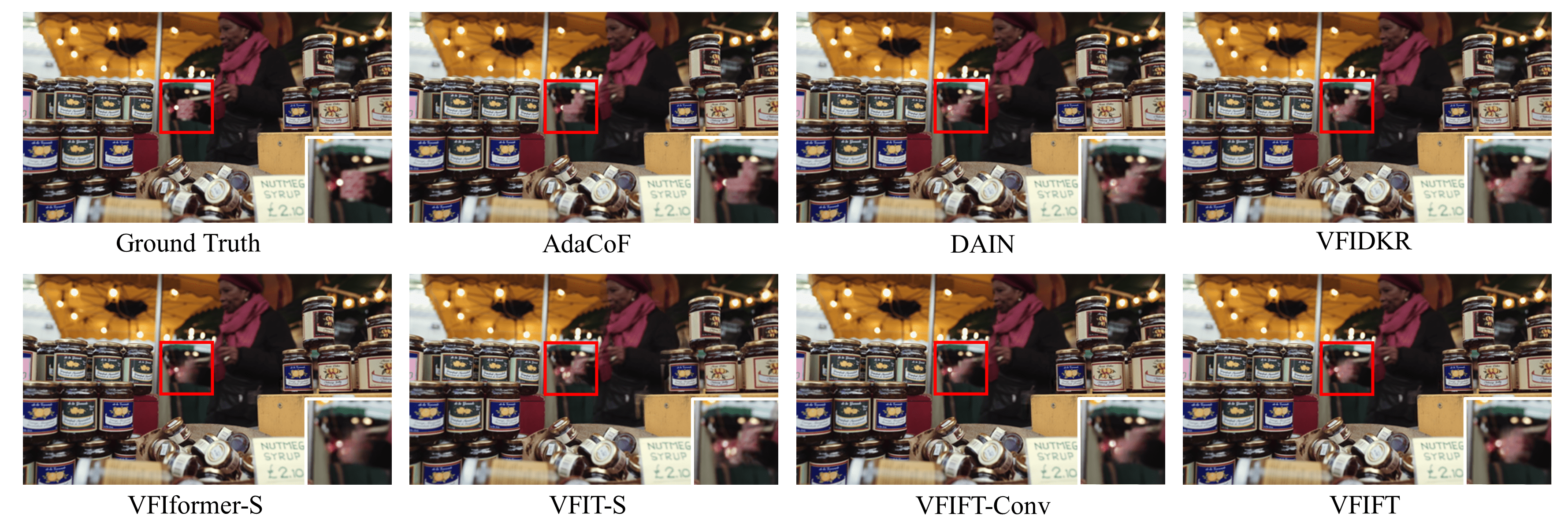}
		
		\vspace{-1mm}
	\end{figure*}

	\begin{figure*}[t]
		\centering
		\vspace{-2mm}
		\includegraphics[width=0.9 \linewidth]{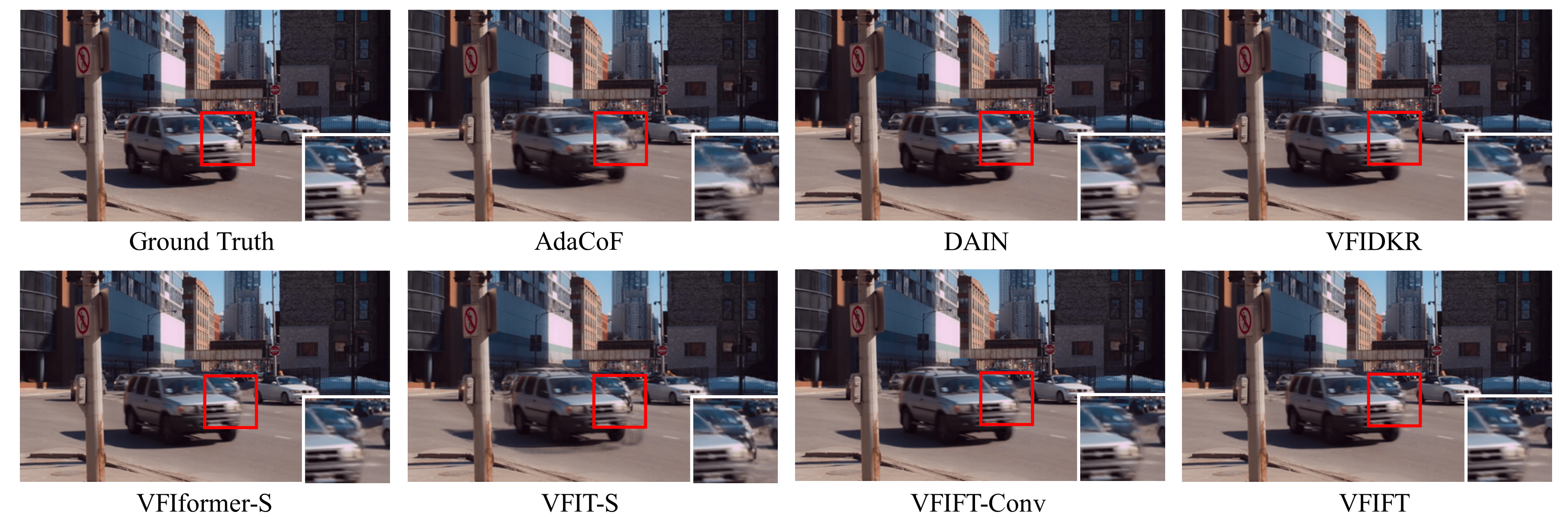}
		
	\end{figure*}

	\begin{figure*}[t]
		\centering
		\vspace{-2mm}
		\includegraphics[width=0.9 \linewidth]{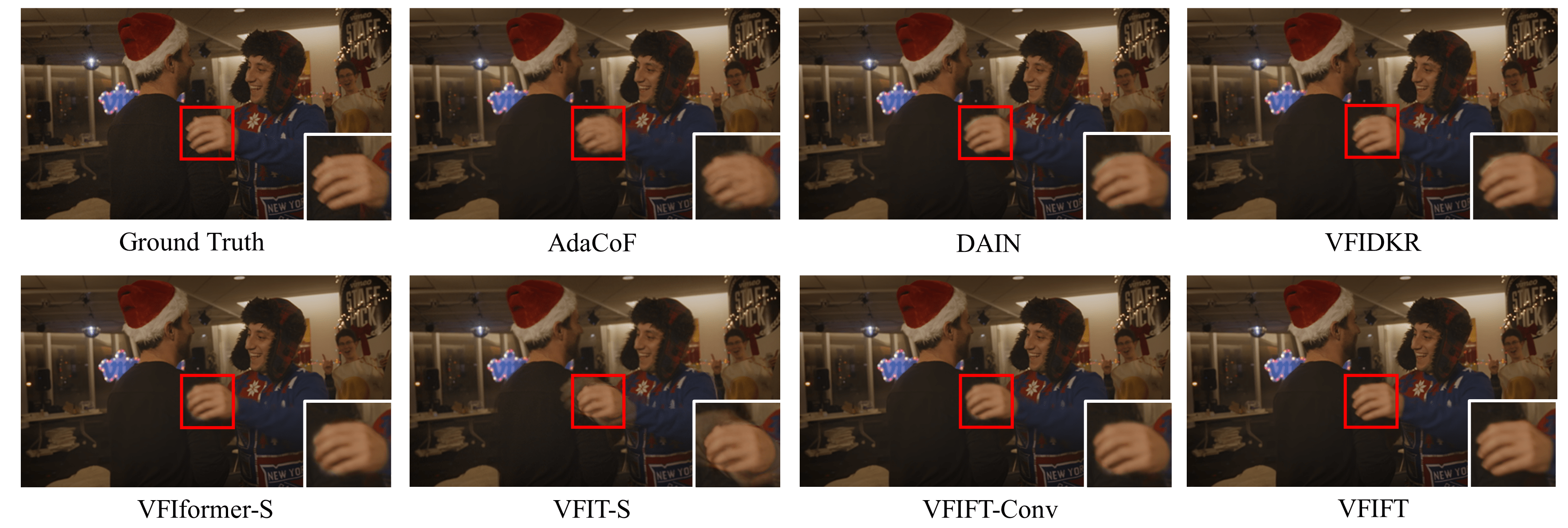}
		
	\end{figure*}

	\begin{figure*}[t]
	\centering
	\vspace{-5mm}
	\includegraphics[width=0.9 \linewidth]{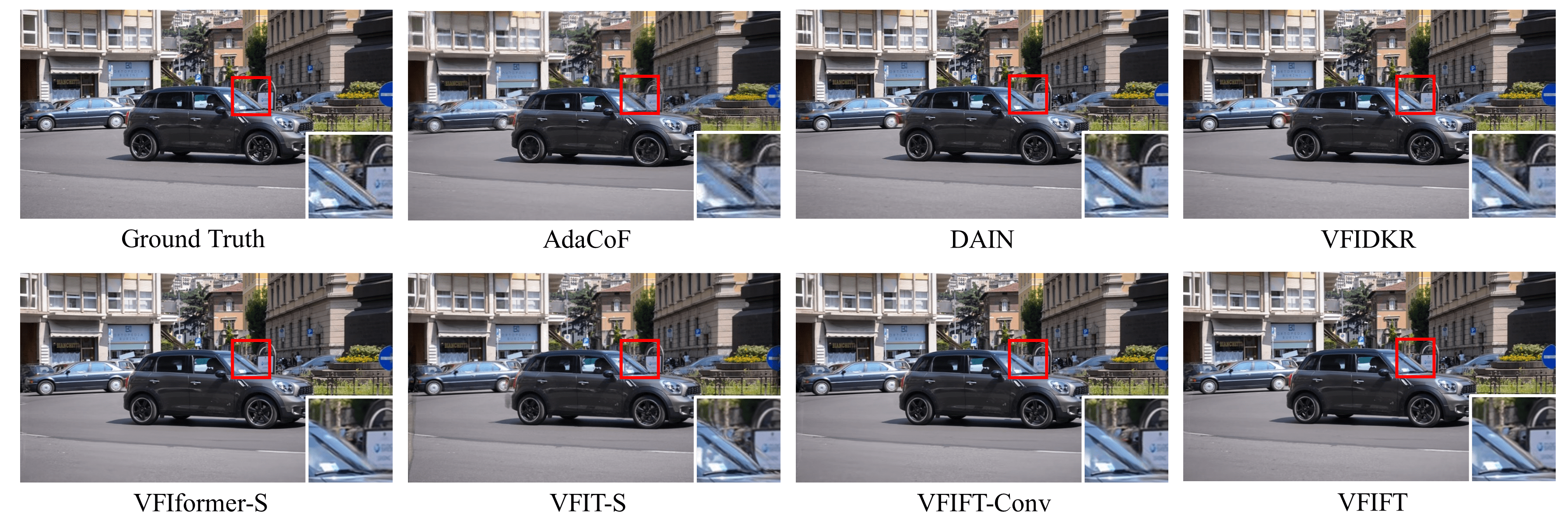}
	
	\end{figure*}
	
	\begin{figure*}[t]
		\centering
		\vspace{-2mm}
		\includegraphics[width=0.9 \linewidth]{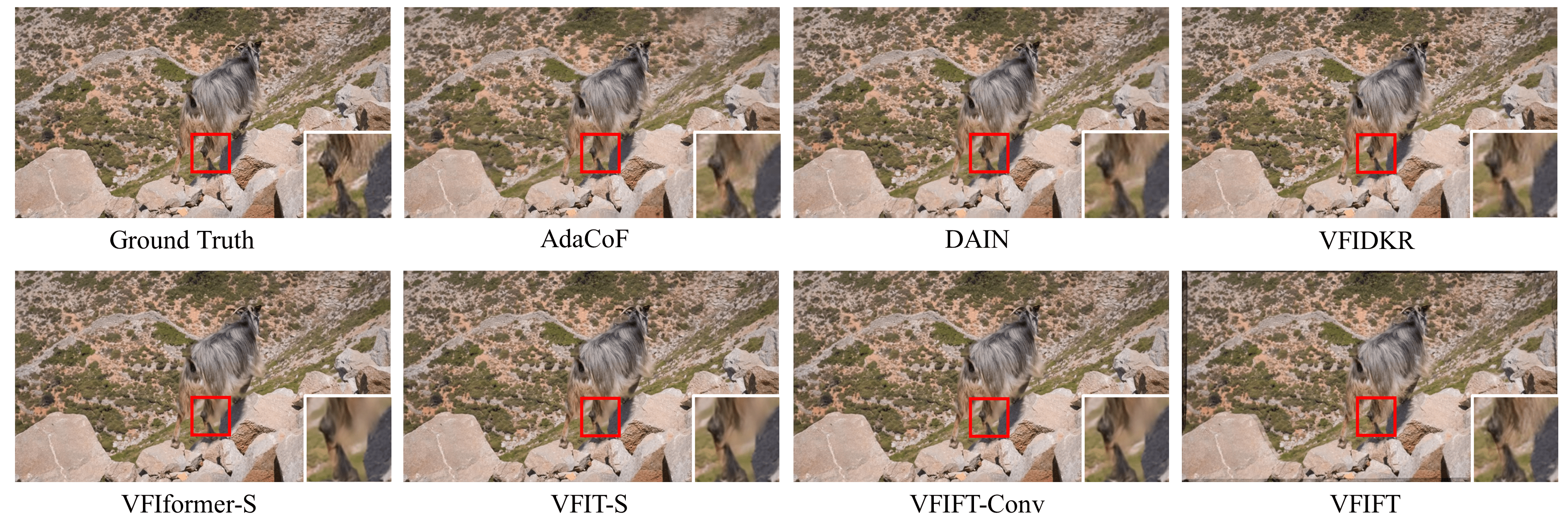}
		
	\end{figure*}
	
	\begin{figure*}[t]
		\centering
		\vspace{-2mm}
		\includegraphics[width=0.9 \linewidth]{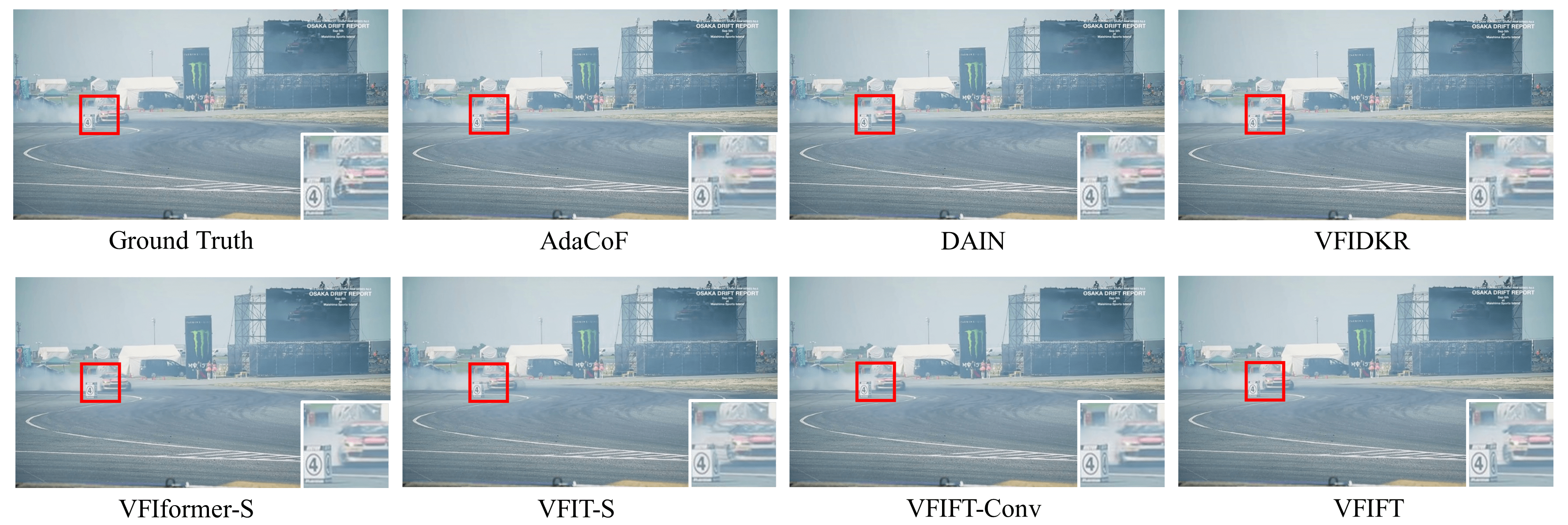}
		
	\end{figure*}
	
\end{document}